\documentclass[letterpaper, 10 pt, conference]{ieeeconf}
\IEEEoverridecommandlockouts

\usepackage{cite}
\usepackage{amsmath,amssymb,amsfonts}
\usepackage{algorithmic}
\usepackage{graphicx}

\def\BibTeX{{\rm B\kern-.05em{\sc i\kern-.025em b}\kern-.08em
    T\kern-.1667em\lower.7ex\hbox{E}\kern-.125emX}}

\title{From Unstable Contacts to Stable Control: A Deep Learning Paradigm for HD-sEMG in Neurorobotics}
\author{Eion Tyacke, Kunal Gupta, Jay Patel, Raghav Katoch, S. Farokh Atashzar,~\IEEEmembership{Senior Member,~IEEE}
\thanks{E. Tyacke is with Carnegie Mellon University (the work was conceptualized when Tyacke was at New York University). K. Gupta, J. Patel, R. Katoch, and S. Farokh Atashzar are with New York University. Atashzar is the corresponding author. {\tt\footnotesize f.atashzar@nyu.edu}.}
}

\begin{document}
\maketitle

\section{abstract}
\textbf{In the past decade, there has been significant advancement in designing wearable neural interfaces for controlling neurorobotic systems, particularly bionic limbs. These interfaces function by decoding signals captured non-invasively from the skin's surface. Portable high-density surface electromyography (HD-sEMG) modules combined with deep learning decoding have attracted interest by achieving excellent gesture prediction and myoelectric control of prosthetic systems and neurorobots. However, factors like pixel-shape electrode size and unstable skin contact make HD-sEMG susceptible to pixel electrode drops. The sparse electrode-skin disconnections rooted in issues such as low adhesion, sweating, hair blockage, and skin stretch challenge the reliability and scalability of these modules as the perception unit for neurorobotic systems. This paper proposes a novel deep-learning model providing resiliency for HD-sEMG modules, which can be used in the wearable interfaces of neurorobots. The proposed 3D Dilated Efficient CapsNet model trains on an augmented input space to computationally  `force' the network to learn channel dropout variations and thus learn robustness to channel dropout. The proposed framework maintained high performance under a sensor dropout reliability study conducted. Results show conventional models' performance significantly degrades with dropout and is recovered using the proposed architecture and the training paradigm.}


 \section{Introduction}

Portable high-density electromyography (HD-sEMG) signal recording has emerged as a promising technology for enhancing wearable neural interfaces and accurately predicting user motor intention, especially for applications like controlling powered prosthesis and bionic limbs \cite{controling_prothetics}. HD-sEMG provides higher spatial resolution by using closely spaced electrodes over a targeted muscle area\cite{highRes_sEMG}. When combined with deep learning decoding architectures like convolutional neural networks (CNNs) and recurrent neural networks (RNNs), the neural interfaces have shown high accuracy ($>\hspace{-0.1cm}95\%$) on gesture classification needed for the control of upper-limb prosthetic devices \cite{highAcc_95, Transformer_based}. 

Of the various deep learning models, Convolutional Neural Network (CNN) and Recurrent Neural Network (RNN) are most frequently used in this context \cite{CNN_RNN_pop,CNN_RNN_pop2,CNN_RNN_pop3}. While CNNs are typically used with a focus on spatial features, RNNs are motivated by encapsulating the temporal dynamics of the signal; both prove beneficial in improving gesture classification accuracy. The hybridization of CNN and RNN has also been explored by recent literature, leading to further improvement in performance by drawing on the benefits of both \cite{CNN_RNN_pop3,CNN_RNN_pop,CNN_RNN_Hybrid}. 

Despite the ability of deep learning models to decode HD-sEMG signal space, their lack of robustness to real-world conditions has prevented large-scale adoption \cite{improvingRobustnessToImproveusage}. Current sEMG is inhibited from mass adoption due to poor spatial resolution of decoding, sensitivity to electrode shift, sensitivity to orientation of placement, unpredictable signal variability, and channel dropout 
\cite{robustnessOrientation,intro_drawback_sEMG,intro_shortcoming_prosthesis,Electrode_Shift_Atashzar,channel_Shift2,intro_drawback_sEMG2,channelDrop}. While there have been some recent attempts at combating HD-sEMG's shortcomings, notably electrode shift\cite{Electrode_Shift_Atashzar}, addressing channel dropout is still an open and important problem.

This paper proposes a deep-learning model to process high-density electromyographic signals for controlling neurorobotic systems. Our model aims to be resilient to common HD-sEMG challenges like sensor dropout and electrode-skin disconnections by effectively utilizing residual data to ensure optimal performance. The model is also designed to achieve high classification accuracy regardless of electrodes shifting over time. This is important because in real-world applications, EMG sensors gradually deviate from their initial calibrated skin positions, leading to substantial accuracy loss with classic decoding neural networks \cite{misplacment_effectSEMG}. Creating a model that maintains performance despite electrode shifts would eliminate the need for frequent sensor reapplication or recalibration.

To capture vital features of  HD-sEMG data, we propose the 3D Dilated Efficient CapsNet (3D CapsNet) neural network model, which we train using the specially augmented input space designed in this paper. CapsNet variants classify objects independent of scale, rotation, and other linear transformations while still drawing information from spacial relationships \cite{orientation_capsnetUse}. This makes the architecture ideal for addressing the altered spacial relationships but maintaining global context caused by channel dropout. We have recently shown that the 2D CapsNet significantly outperforms the standard CNN, MLP, and RNN-CNN hybrid model in gesture classification on transient sEMG \cite{CapsNet}. The proposed augmentation forces the neural network to learn the variations in the input space caused by hundreds of variations of channel dropout. However, due to the logistical complexity of large-scale data collection from human subjects, it would not be feasible to create an augmented input space by conducting experiments using hundreds of HD-sEMG electrodes. Instead, we proposed generating a specially-designed synthetic data augmentation scheme through which the input space will be altered with disturbed synthetic examples made from undisturbed experimental data. This is to force the network to approach better generalizability by being less reliant on the location and dependability of sensors. Augmenting the data will force the model to rely on global channel context to properly classify gestures. 

The results showed that the proposed model, trained based on this specialized augmentation, can maintain significant performance under various complex conditions with signal dropout percentages ranging from 10\% to 75\%, demonstrating robustness and resiliency. Without implementation of the proposed diverse augmentation, a performance drop of $> 37\%$ with only 10\% channels dropped was observed. These results highlight the reliance of conventional deep learning EMG decoding models on precise electrode position/location and that without training for robustness, performance degrades rapidly with dropout.  We evaluated model performance on data from 19 subjects using a publicly available upper limb sEMG database for benchmarking.

To the best of our knowledge, this is the first work to propose a deep learning model that specifically investigated the feasibility of addressing HD-sEMG decoding robustness to channel dropout for gesture prediction. This technique could help improve adoption of HD-sEMG interfaces in real-world prosthetic and robotic systems. In addition to the above-mentioned main contribution, it should be noted that in this paper and to boost translational power, we focused on the transient sEMG, which is a more challenging problem to address compared to the steady-state signal which has been typically used in the literature \cite{steady_state1,steady_state2}. The use of transient sEMG would allow us to enhance the agility in response for real-world scenarios, reduce delay, and make the controls for prosthetic limbs more intuitive\cite{transient1}.
The rest of the paper is organized as follows: Section II provides background on the dataset, data preprocessing, data augmentations, and training strategies; Section III describes the proposed model architectures; Section IV presents the experimental setup, results, and discussion; and Section V concludes the paper.

\section{Materials and Method}
\subsection{Data Acquisition:}\label{A}
he HD-sEMG dataset comprises 20 participants who each performed 65 distinct hand gestures over five repetitions. Two 64-electrode grids positioned over the extensor and flexor forearm muscles recorded the HD-sEMG signals, sampled at 2048Hz using the Quattrocento biomedical amplifier system. The raw signals were hardware band-pass filtered at 10Hz to 900Hz, and the 50Hz power line frequency removed with a zero-phase 3rd order band-pass Butterworth filter with a 4Hz width \cite{65_database}. As subject 5's data was corrupted, this paper utilized the data from the remaining 19 subjects.

\subsection{Data Preprocessing:}\label{B}
In this paper, the time window of 1.2 seconds is included from each repetition, which primarily contains the transient and more dynamic phase of the contraction (more details in \cite{Electrode_Shift_Atashzar, Deep_Heterogeneous}). This is done to enhance the system's agility in responding to user intentions. The HD-sEMG signals then undergo a three-part transformation before windowing. The data passes through a 4th-order Butterworth band-pass filter from 10Hz to 500Hz, further reducing possible noise or interference and ensuring the signal primarily contains relevant muscle activation information. Z-score normalization then scales the data to zero mean and unit standard deviation. Repetitions 1, 3, and 4 constitute the training data, while repetitions 2 and 5 are the testing data. The data is then rectified, converting all negative values to positive. A sliding prediction window with a length of 200ms and increment of 10ms (95\% overlap) is applied.

\subsection{Data Augmentation:}\label{C}
\begin{figure}[!hbt]
    \centering
    \includegraphics[width=0.4\textwidth]{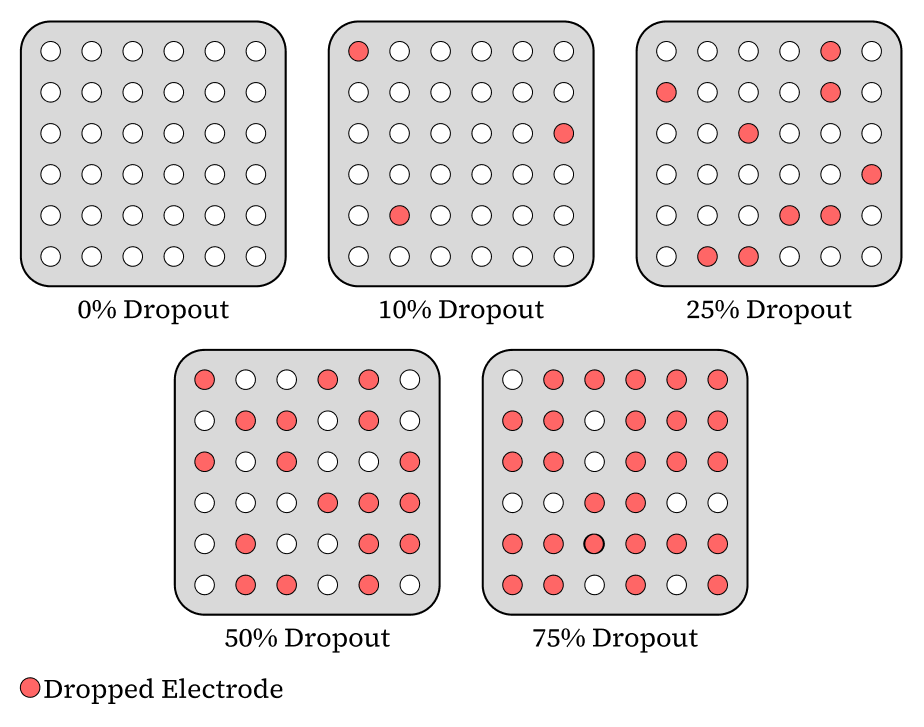}
    \caption{5 HD-sEMG modules demonstrating the effect of the channel drop augmentation. A white circle indicates a functional sensor while a red circle indicates a synthetically dropped sensor. The number of dropped sensors is in accordance to the labeled dropout rate.}
    \label{Dropout_Mask_Image}
    \vspace{-0.3cm}
\end{figure}

For data augmentation, randomized channel dropout is applied to accurately simulate real-world dropped signals. For each subject, when a channel is dropped, its value is set to zero along the entire recording. Channel dropout is defined by masks. A mask is a $6\times6$ Boolean array with each value corresponding to each electrode, with a true value indicating a channel should be dropped. Masks are generated according to a dropout rate. For example, if there are 36 electrodes and a mask is generated with a 0.5 dropout rate, 18 channels will be dropped. Each "ring" of the $6\times6$ mask has an equal proportion of the ring dropped out. For example, with a 0.75 dropout rate, the outer ring containing 20 electrodes will have 15 electrodes randomly dropped, while the second inner ring containing 12 electrodes will have 9 electrodes randomly dropped. In this paper, models are trained on 5 dropout rates: 0\%, 10\%, 25\%, 50\%, 75\%. Figure \ref{Dropout_Mask_Image} visualises the 5 differnt channel dropout augmentations. For each rate, 6 masks were generated and individually applied to the dataset, then combined with a duplicate of the clean data, effectively increasing the training data seven-fold. We also trained a model on all rates, combining the data of every augmented input space. Multiple masks were used because when trained on a single mask, the model only recognized the dropped pattern of that mask, not improving robustness. This augmentation markedly increased variety in the dataset and overall algorithm performance, mitigating overfitting and improving accuracy compared to models trained on a single mask. In addition, synthetic electrode shift is also applied to simulate slight misplacement of the grids without requiring extensive recollected data with varying placements. Beginning with the two full $8\times8$ grids, a kernel window of $1\times6\times6\times1$ is selected and the patches extracted with stride one. This procedure is synonymous with the combination of one-step and two-step shifting described by \cite{Electrode_Shift_Atashzar}.

\section{Model Structure}
The validated architecture in this paper is named  3D Dilated Efficient CapsNet and is compared with the performance of 3D CNN as the conventional approach.
 
\subsection{3D CNN}\label{A}
The 3D CNN model discussed in this paper follows the settings from \cite{Electrode_Shift_Atashzar}. The model consists of a block with five 3D convolutional layers and a classifier with two dense layers. The initial four convolutional layers have a kernel size of (100, 2, 2) and filter counts of 8, 16, 32, and 64, respectively. The last convolutional layer has a (4, 2, 2) kernel size with 128 filters. The dense classifier has hidden layers of sizes 128 and 96.

\begin{figure*}[hbt!]
    \centering
    \includegraphics[width=0.95\textwidth]
    {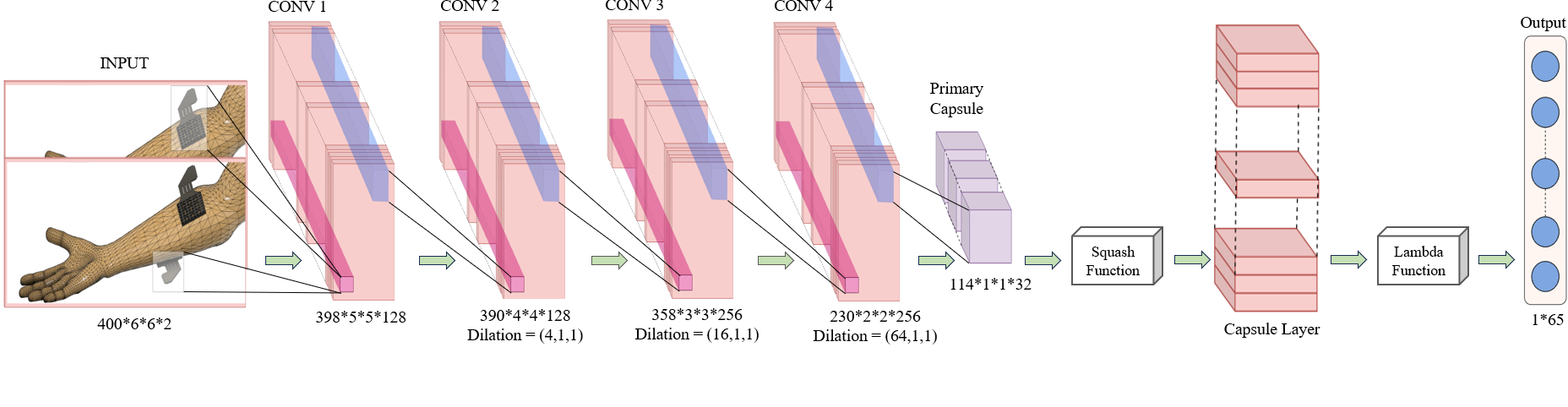}
    \vspace{-0.2cm}
    \caption{An illustration of the proposed 3D Dilated Efficient CapsNet architecture. There are four 3D Convolutional layers (Conv). The Capsule block has a Primary Conv followed by a squash function, a Capsule layer to perform routing-by-agreement, and a Lambda Function to generate the output confidence for each of the 65 gestures.}\vspace{-0.3cm}
    \label{F-model}
\end{figure*}

\subsection{3D Dilated Efficient CapsNet}\label{C}
The 3D CapsNet architecture (see Fig. \ref{F-model}), inspired by \cite{CapsNet}, enhances the original Dilated Eff-Caps model for the utilization of 3D data through the conversion of 2D convolutional layers to 3D convolutional layers. In general, in capsular networks, the model classifies objects through collections of neurons called capsules. Each capsule outputs a vector whose length describes the probability of a feature existing and whose orientation describes the feature's pose. High-probability capsules then predict the output of capsules in the subsequent layer in a process known as routing by agreement. When clusters of capsules agree on a capsule's output in the subsequent layer, that capsule is activated. The process is repeated on each layer using activated capsules until a steady state is reached. This architecture organizes a feature hierarchy not reliant on spatial information, making it excellent at classifying gestures despite augmented input spaces. Within the convolutional block of the model, the first two convolutional layers have 128 features each, and the last two have 256 features; each layer possesses a kernel size of (3, 2, 2) and is followed by a dropout layer with a dropout rate of 0.5 and a batch normalization layer. To increase the receptive field along the temporal dimension for successive layers without increasing the convolutional kernel size, we employ an increasing dilation rate from convolutional layers two through four with rates of (4, 1, 1), (16, 1, 1), and (64, 1, 1). Within the capsule block, both the primary convolution and the capsule layer utilized a capsule dimension of eight. Additionally, the primary convolution layer has four channels, a kernel size of (3, 1, 1), and a stride of two.

\section{Experiment and Results}
To evaluate the 3D CNN and CapsNet performances, each model was tested at 0\%, 10\%, 25\%, 50\%, and 75\% channel dropout rates. Each test assessed a model's average performance across 30 randomly selected drop channel patterns. These randomly selected test masks are distinct from those used in the processes of generating the augmented input space for training and are consistent between each model evaluation. Fig. \ref{CNN_line_plot} and \ref{CapsNet_line_plot} display their respective base model and its variants. The x-axis shows rising dropout percentages, and the y-axis shows the 30-mask average accuracy of each model.

To better evaluate our models’ robustness, we used the two-sided Mann-Whitney-Wilcoxon test for statistical significance testing. This test compared the performance distributions between model variants tested with 0\% channel dropout and each increasing level of channel dropout. The Bonferroni correction was applied to provide a more conservative estimate when conducting multiple tests, reducing the risk of making a Type I error. In this paper, significance markers ns, *, **, ***, and **** correspond to p-values of 1-0.05, 0.05-0.01, 0.01-0.001, 0.001-0.0001, and $<0.0001$ respectively. The results of this testing can be seen in Tables \ref{CNN_stats_table} and \ref{CapsNet_stats_table}.

\subsection{Performance evaluation for the 3D CNN architecture}
\begin{figure}[!hbt]
    \centering
    \includegraphics[width=0.38\textwidth]{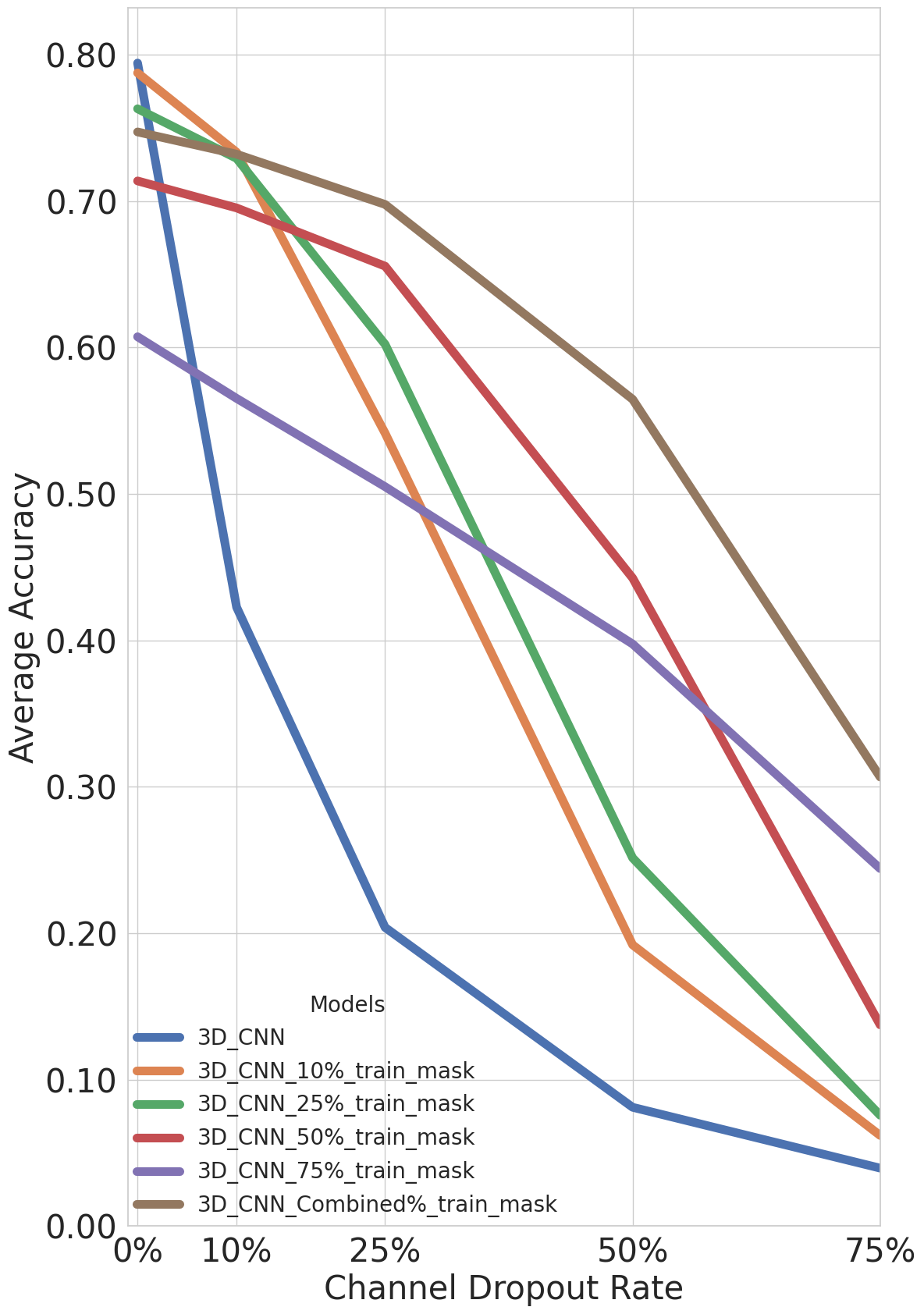}
    \caption{3D CNN average accuracy results for each level of augmentation evaluated on 30 randomly assigned dropped channel configurations at each dropout rate. X-axis is the dropout rate and different colors show different augmentation strategy.}
    \label{CNN_line_plot}
\end{figure}

\begin{figure}[hbt!]
    \centering
    \includegraphics[width=0.38\textwidth]{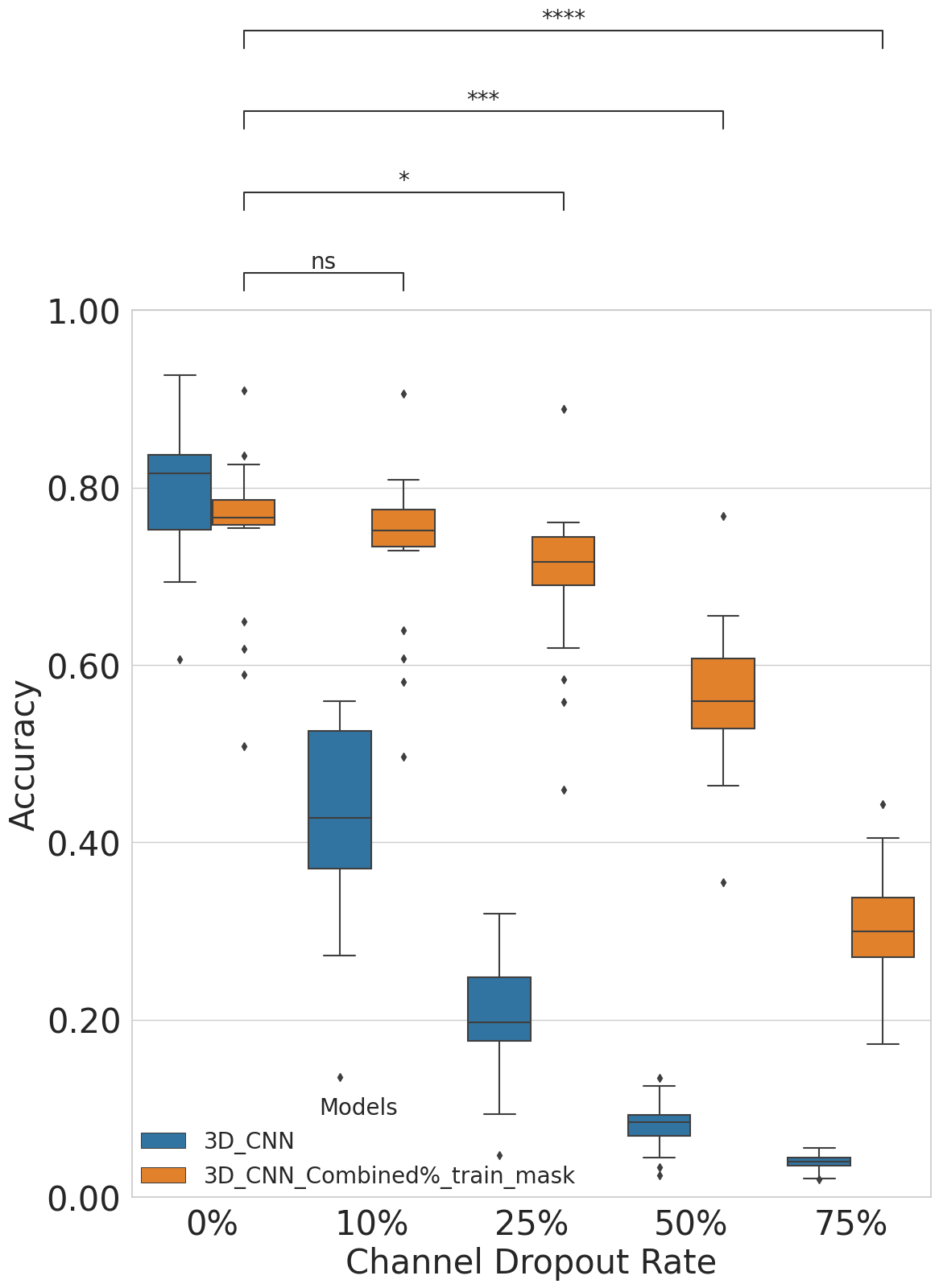}
    \caption{3D CNN results at each dropout level for training on the combination of all channel dropout data augmented strategies for each level}
    \label{CNN_box_plot}
\end{figure}

We compared a 3D CNN trained with 0\% dropout (baseline) to the same model trained on five augmented input spaces. Results are shown in Fig. \ref{CNN_line_plot} and \ref{CNN_box_plot} and are summarized in Tables. \ref {CNN_acc_table}, and \ref{CNN_stats_table}. As illustrated in Fig. \ref{CNN_line_plot}, the augmented input spaces (shown by different colors) are tested for channel dropout rates ranging from 10\% to 75\% (x-axis of the figure). The figure demonstrates that our augmented training approach improves the robustness of the gesture classifier. This enhancement is evident from the superior average accuracy rates achieved by all non-baseline models under dropout conditions. 
As the dropout rate in the augmented training set increases (represented by different color codes in Fig. \ref{CNN_line_plot} and \ref{CNN_box_plot}) from 0\% to 75\%, the impact of randomized dropout intensity (represented by the x-axis of Fig. \ref{CNN_line_plot} and \ref{CNN_box_plot}) reduces, resulting in a boost in the model's average accuracy. The increasing trend of 'no significant difference' across the rows of Table \ref{CNN_stats_table} indicates enhanced consistency of model performance and thus robustness to dropout. Both Fig. \ref{CNN_line_plot} and \ref{CNN_box_plot} emphasize this, with non-baseline models displaying higher averages and smaller slopes (which means less variability caused by dropout). Notably, the model trained with the combined training mask stands out with the highest accuracy and the highest robustness to randomized dropouts shown, underscoring the performance of the proposed augmentation. \vspace{-0.3cm}

\begin{table}[hbt!]
  \begin{center}
    \caption{Results of Channel Dropout Rates on 3D CNN \hspace{1.3cm}(Average Accuracy)}
    \label{CNN_acc_table}
        \begin{tabular}{|c|c|c|c|c|c|}
            \hline $\begin{array}{l}\textbf{\hspace{0.3cm}Dropout} \\
            \textbf{\hspace{0.4cm}Rate$\rightarrow$}\end{array}$ & \textbf{0\%} & \textbf{10\%} & \textbf{25\%} & \textbf{50\%} & \textbf{75\%} \\
            \hline
            \textbf{0\% Augmentation} & 0.794 & 0.423 & 0.204 & 0.081 & 0.040 \\
            \textbf{10\% Augmentation} & 0.787 & 0.733 & 0.541 & 0.192 & 0.062 \\
            \textbf{25\% Augmentation} & 0.763 & 0.729 & 0.602 & 0.251 & 0.076 \\
            \textbf{50\% Augmentation} & 0.714 & 0.695 & 0.655 & 0.442 & 0.137 \\
            \textbf{75\% Augmentation} & 0.607 & 0.565 & 0.505 & 0.397 & 0.244 \\
            \textbf{Combined} & 0.747 & 0.732 & 0.698 & 0.565 & 0.307 \\
            \hline
        \end{tabular}
   \end{center}
   \vspace{-0.6cm}
\end{table}

\begin{table}[hbt!]
  \begin{center}
    \caption{Results of Channel Dropout Rates on 3D CNN \hspace{1cm} (Statistical Significance)}
    \label{CNN_stats_table}
        \begin{tabular}{|c|c|c|c|c|c|}
            \hline $\begin{array}{l}\textbf{\hspace{0.3cm}Dropout} \\
            \textbf{\hspace{0.4cm}Rate$\rightarrow$}\end{array}$ & \textbf{0\%} & \textbf{10\%} & \textbf{25\%} & \textbf{50\%} & \textbf{75\%} \\
            \hline
            \textbf{0\% Augmentation} & - & **** & **** & **** & **** \\
            \textbf{10\% Augmentation} & - & * & **** & **** & **** \\
            \textbf{25\% Augmentation} & - & ns & **** & **** & **** \\
            \textbf{50\% Augmentation} & - & ns & ns & **** & **** \\
            \textbf{75\% Augmentation} & - & ns & ** & **** & **** \\
            \textbf{Combined} & - & ns & * & *** & **** \\
            \hline
        \end{tabular}
   \end{center}
   \vspace{-0.5cm}
\end{table}

\subsection{Performance evaluation for the 3D Dilated Efficient CapsNet architecture}

\begin{figure}[hbt!]
    \centering
    \includegraphics[width=0.38\textwidth]{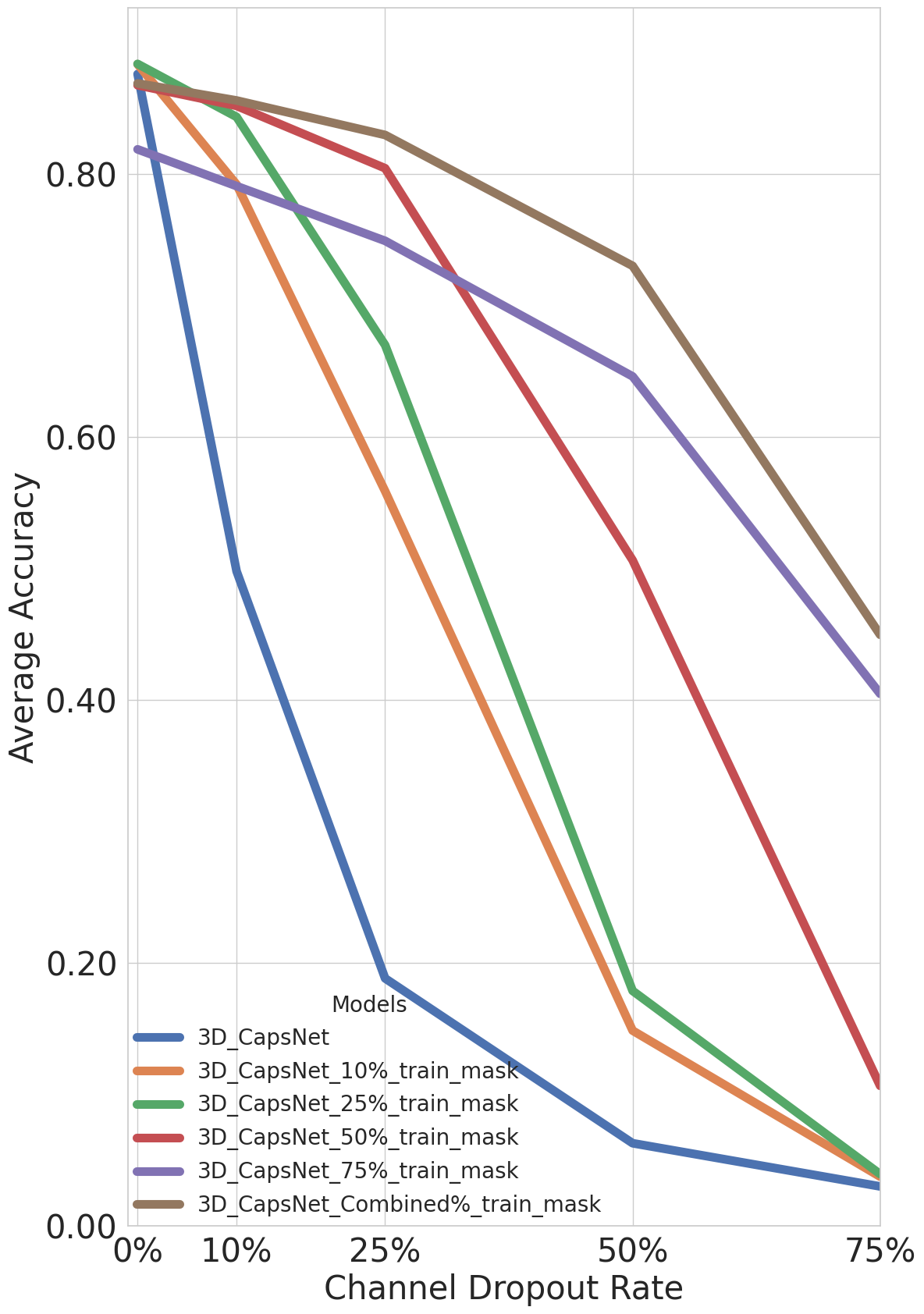}
    \caption{3D CapsNet average accuracy results for each level of augmentation evaluated on 30 randomly assigned dropped channel configurations at each dropout rate} \vspace{-0.5cm}
    \label{CapsNet_line_plot}
\end{figure}

\begin{figure}[hbt!]
    \centering
    \includegraphics[width=0.38\textwidth]{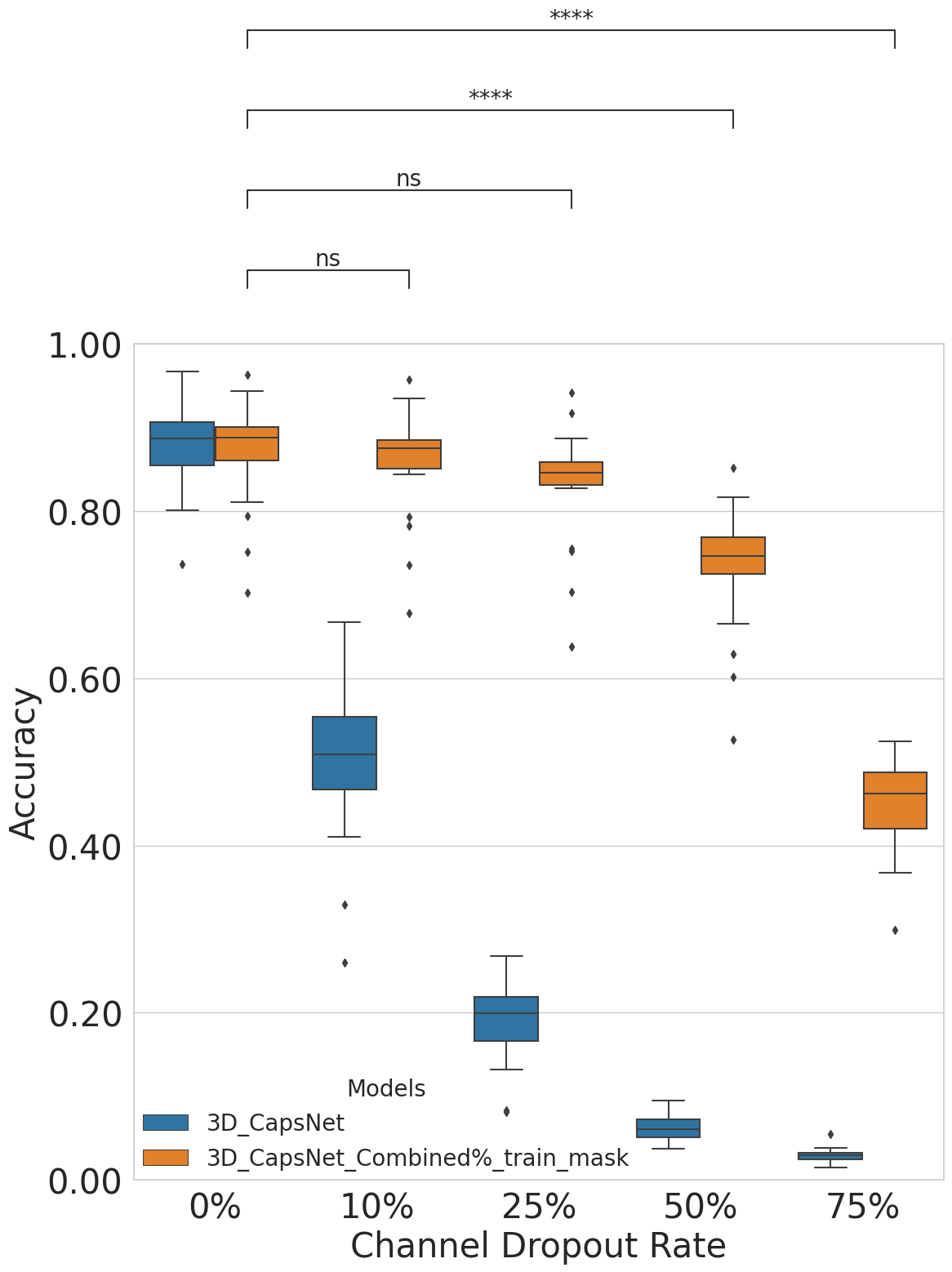}
    \caption{3D CapsNet results at each dropout level for training on the combination of channel dropout data augmented for each level}\vspace{-0.6cm}
    \label{CapsNet_box_plot}
\end{figure}

Similarly to the 3D CNN, our proposed CapsNet model was tested against different channel dropout levels, with and without training on augmented datasets. The results, shown in Fig. \ref{CapsNet_line_plot} and \ref{CapsNet_box_plot}, reveal that the CapsNet models boosted by the augmented training on channel dropout outperform the baseline model (when no augmentation is conducted). When the proposed CapsNet is trained on the combination of all augmentation data (Table \ref{CapsNet_stats_table} and Fig. \ref{CapsNet_box_plot}), consistent model performance was observed with no statistically significant difference when dropping 25\%, 10\% and 0\% of the channels; and the model robustly maintains the performance of ~83\%, resembling robustness, making it the best-performing model among all the CapsNets and 3D CNNs evaluated in this study. Conducting a stronger stress test, the performance of the proposed model remains above ~70\% even in the presence of 50\% dropout, when the baseline model can only maintain the performance of $\approx6\%$ (see Table III). The results highlight the resiliency of the proposed model to even severe cases of drop out highlighting the importance of the proposed synthetic input augmentation for practical application.

\begin{table}[hbt!]
  \begin{center}
    \caption{Results of Channel Dropout Rates on 3D CapsNet \hspace{1cm}(Average Accuracy)}
    \label{CapsNet_acc_table}
        \vspace{-.2cm}
        \begin{tabular}{|c|c|c|c|c|c|}
            \hline $\begin{array}{l}\textbf{\hspace{0.3cm}Dropout } \\
            \textbf{ \hspace{0.4cm}Rate$\rightarrow$}\end{array}$ & \textbf{0\%} & \textbf{10\%} & \textbf{25\%} & \textbf{50\%} & \textbf{75\%} \\
            \hline
            \textbf{0\% Augmentation} & 0.876 & 0.498 & 0.188 & 0.063 & 0.030 \\
            \textbf{10\% Augmentation} & 0.884 & 0.792 & 0.559 & 0.149 & 0.038 \\
            \textbf{25\% Augmentation} & 0.884 & 0.844 & 0.670 & 0.179 & 0.039 \\
            \textbf{50\% Augmentation} & 0.868 & 0.852 & 0.805 & 0.506 & 0.107 \\
            \textbf{75\% Augmentation} & 0.819 & 0.791 & 0.749 & 0.646 & 0.405 \\
            \textbf{Combined} & 0.869 & 0.856 & 0.830 & 0.730 & 0.450 \\
            \hline
        \end{tabular}
   \end{center}
\end{table}

\begin{table}[hbt!] \vspace{0.5cm}
  \begin{center}
    \caption{Results of Channel Dropout Rates on 3D CapsNet \hspace{1cm} (Statistical Significance)}
    \label{CapsNet_stats_table}
        \vspace{-.2cm}
        \begin{tabular}{|c|c|c|c|c|c|}
            \hline $\begin{array}{l}\textbf{\hspace{0.3cm}Dropout} \\
            \textbf{\hspace{0.4cm}Rate$\rightarrow$}\end{array}$ & \textbf{0\%} & \textbf{10\%} & \textbf{25\%} & \textbf{50\%} & \textbf{75\%} \\
            \hline
            \textbf{0\% Augmentation} & - & **** & **** & **** & **** \\
            \textbf{10\% Augmentation} & - & *** & **** & **** & **** \\
            \textbf{25\% Augmentation} & - & ns & **** & **** & **** \\
            \textbf{50\% Augmentation} & - & ns & ** & **** & **** \\
            \textbf{75\% Augmentation} & - & ns & * & **** & **** \\
            \textbf{Combined} & - & ns & ns & **** & **** \\
            \hline
        \end{tabular}
   \end{center}
   \vspace{-0.5cm}
\end{table}

Figures \ref{CNN_box_plot} and \ref{CapsNet_box_plot} demonstrate that models trained on the combined augmented dataset maintain significantly high accuracy and exhibit smaller interquartile ranges up to a 25\% experimental channel dropout rate, compared to their base model. The results in Fig. \ref{CapsNet_line_plot} and Table \ref{CapsNet_acc_table} suggest that models trained with augmentation by higher dropout rates tend to outperform those trained with augmentation by lower dropout rates, even when evaluated on data with a lower number of channels dropped. Notably, the 3D CapsNet model, when trained through augmentation by 50\% dropout augmentation strategy, outperformed the model augmented by 10\% dropout strategy and 25\% dropout strategy for all experimental ratios of sensor dropout (including the low and high rates). As expected when combining all the augmentation strategies for training the CapsNet due to the large size of input space it requires much higher computational support for training. Thus based on the results, it seems that training the model using an augmentation strategy of 50\% dropout emerges as an efficient and less resource-intensive option with high performance. The model trained by 50\% dropout augmentation can simulate a range of dropout scenarios by learning diverse patterns between small and sparse clusters of electrodes rather than relying on a limited number of large, dense patches. This highlights the significance of learning features at various scales, as the smaller and sparser features learned at the 50\% augmentation level contribute to better generalization. It should be highlighted that training to combat higher dropout rates is a tradeoff between computing resources and accuracy. As the channel dropout rate increases, the possible combinations of channels that remain active, or ‘live’, also increases. This means that to effectively train the model at higher dropout rates, the number of masks used to augment the training data should be increased. This would expose the model to a wider range of potential configurations of ‘live’ channels. However, it can increase the computational cost during the training.

\section{Acknowledgement}
The authors would like to acknowledge the help of Ananya Shah in the project.
\section{Conclusion}

This paper presents a specialized data augmentation methodology to increase the performance of high-density sEMG-based decoding with ultimate application in neurorobotics. The goal is to recover robustness in the presence of sensor pixel dropouts, which can significantly challenge the performance of the interfaces in practice. Applying the proposed data augmentation methodology on the majority transient phase of HD-sEMG signal and training the proposed Capsular Neural Network model has shown the viability of the sustained performance of such systems in a degraded setting, achieving an accuracy of 83\% when 0-25\% channel dropout was present for 65 gesture prediction. The model kept the performance above 70\% in the presence of severe dropout (i.e., 50\%) when the baseline model could only provide $\approx6\%$ performance due to the complexity of the task. In other words using the proposed framework we have shown that high-density electromyography provides a very rich source of information that can secure high performance if the model is trained in a way that do not rely on spatial locations of the information source across the targeted muscle. Addressing the drop-out issue through the proposed robust system opens the door to wider acceptance of peripheral wearable devices for control of neurorobots.

\bibliographystyle{ieeetr}
\bibliography{ref}

\end{document}